\documentclass{article}

\usepackage{arxiv}
\usepackage[utf8]{inputenc} 
\usepackage[T1]{fontenc}    
\usepackage{hyperref}       
\usepackage{url}            
\usepackage{booktabs}       
\usepackage{nicefrac}       
\usepackage{microtype}      
\usepackage{multirow}
\usepackage{tabularx}
\usepackage{ltablex}
\usepackage{rotating}
\usepackage{amsmath,amsfonts}
\usepackage[ruled,vlined]{algorithm2e}

\title{Comparative Study Between Distance Measures On Supervised Optimum-Path Forest Classification}

\author{
  Gustavo H. de Rosa, Mateus Roder, João P. Papa \\
  Department of Computing \\
  São Paulo State University \\
  Bauru, São Paulo - Brazil \\
  \texttt{\{gustavo.rosa, mateus.roder, joao.papa\}@unesp.br} \\
}

\begin{document}

\maketitle

\begin{abstract}
Machine Learning has attracted considerable attention throughout the past decade due to its potential to solve far-reaching tasks, such as image classification, object recognition, anomaly detection, and data forecasting. A standard approach to tackle such applications is based on supervised learning, which is assisted by large sets of labeled data and is conducted by the so-called classifiers, such as Logistic Regression, Decision Trees, Random Forests, and Support Vector Machines, among others. An alternative to traditional classifiers is the parameterless Optimum-Path Forest (OPF), which uses a graph-based methodology and a distance measure to create arcs between nodes and hence sets of trees, responsible for conquering the nodes, defining their labels, and shaping the forests. Nevertheless, its performance is strongly associated with an appropriate distance measure, which may vary according to the dataset's nature. Therefore, this work proposes a comparative study over a wide range of distance measures applied to the supervised Optimum-Path Forest classification. The experimental results are conducted using well-known literature datasets and compared across benchmarking classifiers, illustrating OPF's ability to adapt to distinct domains.
\end{abstract}


\section{Introduction}
\label{s.introduction}

The advance of computational power has strengthened humans' capacity to solve Artificial Intelligence (AI) tasks in real-world activities~\cite{Hwang:18}, such as image classification and reconstruction~\cite{Wang:18}, object recognition~\cite{Sukanya:16}, and medical analysis~\cite{Litjens:17}, among others. Moreover, it also facilitated the growth of an essential subarea, denoted as Machine Learning (ML), which fosters autonomous-based algorithms applied to AI tasks.

ML algorithms are often categorized into two distinct types of learning~\cite{Alpaydin:20}: (i) supervised and (ii) unsupervised. The former stands for algorithms, such as Linear Regression (LR)~\cite{Montgomery:12}, Logistic Regression (LogR)~\cite{Hosmer:13}, Decision Trees (DT)~\cite{Safavian:91}, Support Vector Machines (SVM)~\cite{Chang:11}, Optimum-Path Forests (OPF)~\cite{Papa:12}, that learn mathematical functions capable of mapping inputs to outputs based on given examples and are usually applied to classification and regression tasks. The latter algorithms, such as $K$-Means~\cite{Wagstaff:01}, Density-Based Spatial Clustering (DBSCAN)~\cite{Khan:14}, Gaussian Mixture Models (GMM)~\cite{Reynolds:09}, focus on modeling the intrinsic structure of the given data as there is no output to be mapped, and is mostly applied to clustering- and association-based tasks.

One common problem amongst all ML algorithms concerns their hyperparameter selection, which may lead to poor performance and the inability to solve the desired task when wrongly chosen. For instance, LogR can be used with or without regularization penalties and have a tolerance stopping criteria. DT needs a maximum depth parameter that controls the number of nodes' expansions, while SVM supports distinct kernels, affecting the outcome when mistakenly chosen. Furthermore, $K$-Means demands the setting of how many clusters will be used, while DBSCAN needs a maximum distance to specify whether two samples are considered or not in the same neighborhood.

For instance, Yogatama et al.~\cite{Yogatama:14} addressed hyperparameter tuning through a transfer learning approach, where previous datasets' meta-information fit a Gaussian Process to construct a joint response surface and further estimate new hyperparameters. Mantovani et al.~\cite{Mantovani:15} proposed to find suitable SVM hyperparameters through a simple Random Search, which is less expensive than meta-heuristic optimization and Grid Search, and could produce equivalent results. Furthermore, Probst et al.~\cite{Probst:19} presented the tuneRanger package, which automatically tunes RF hyperparameters through a model-based optimization. At the same time, Wu et al.~\cite{Wu:19} used Gaussian Processes to model an optimization problem among Machine Learning models and their hyperparameters, solving it through Bayesian optimization.

On the other hand, the graph-based classifier denoted as Optimum-Path Forest has a supervised version which only depends on a distance measure to create arcs between the graph's nodes and further assemble them into trees. The resulting trees compose the so-called forest and are responsible for conquering new nodes added to the graph, propagating the tree root's label to the conquered nodes. Even though the OPF classifier works in almost any application with its standard distance measure\footnote{Such a distance is a logarithm-based version of the euclidean distance, weighed by a maximum arc value.}, there might be the need to adjust it whenever it is incapable of producing an appropriate result. To the best of the authors' knowledge, few works attempt to incorporate multiple distance spaces in OPF classification~\cite{Silva:12,Mansano:14}, yet it does not provide a thorough comparison amongst various distance measures. Additionally, only a similar work attempts to verify the effects of distance measures in supervised learning; however, in K-Nearest Neighbor classification~\cite{Abu:19}.

This work addresses such a problem by performing a comparative study between $47$ distance measures applied to OPF classifiers in a wide range of applications, ranging from image classification to biomedical analysis. Additionally, we opted to employ several baseline classifiers with distinct sets of hyperparameters, such as LR, LogR, DT, and SVM, to provide a more robust comparison and check whether OPF distance selection is less sensitive than hyperparameter fine-tuning. Therefore, the main contributions of this work are three-fold: (i) to introduce new distance measures in supervised OPF, (ii) to provide a robust comparison amongst distance-selected OPF and hyperparameter-tuned baseline classifiers, and (iii) to fill the lack of research regarding OPF-based classification.

The remainder of this work is organized as follows. Section~\ref{s.background} presents a theoretical background regarding the supervised Optimum-Path Forest, as well as the employed distance measures. Section~\ref{s.methodology} introduces the proposed approach, datasets, and experimental setup while Section~\ref{s.experiments} presents the experimental results and an in-depth discussion about them. Finally, Section~\ref{s.conclusion} states the conclusions and future works.
\section{Theoretical Background}
\label{s.background}

This section presents a more in-depth explanation regarding the Optimum-Path Forest classifier and distance measures.

\subsection{Optimum-Path Forest}
\label{ss.opf}

Papa et al.~\cite{Papa:09} proposed the Optimum-Path Forest, a multi-class graph-based classifier capable of segmenting the feature space without using massive data. Essentially, the OPF aims at constructing a complete graph, where an arc connects every pair of nodes. During its training phase, OPF defines a set of prototype\footnote{Master node that represents a specific class and conquer other nodes.} nodes and let them compete between themselves to conquer the remaining nodes, i.e.,  finding the best path (lowest cost). Afterward, during the testing phase, OPF inserts a new sample in the graph and finds a minimum cost prototype, defining its label.

Let $\mathcal{Z}$ be a dataset composed of training and testing sets denoted as $\mathcal{Z}_1$ and $\mathcal{Z}_2$, respectively. One can define a graph $G = (\mathcal{V},\mathcal{A})$\footnote{Note that one can also define $G_1 = (\mathcal{V}_1,\mathcal{A}_1)$ and $G_2 = (\mathcal{V}_2,\mathcal{A}_2)$ from $\mathcal{Z}_1$ and $\mathcal{Z}_2$, respectively.} which belongs to $\mathcal{Z}$ such that $v(s) \in \mathcal{V}$, where $s$ stands for a sample in dataset $\mathcal{Z}$ and $v(\cdot)$ stands for a feature extraction function. Additionally, let $\mathcal{A}$ be an adjacency relation that connects samples in $\mathcal{V}$, as well as let $d : \mathcal{V} \times \mathcal{V} \rightarrow \Re^+$ be a distance function that weigh edges in $\mathcal{A}$.

\subsubsection{Training Step}
\label{sss.opf_train}

Let $\pi_s$ be a path in $G$ that ends in node $s \in \mathcal{V}$ and let $\langle \pi_s \cdot (s,t) \rangle$ be the nexus between path $\pi_s$ and arc $(s,t) \in \mathcal{A}$. The Optimum-Path Forest classifier aims at establishing a set of prototypes nodes $\mathcal{S} \subseteq \mathcal{V}$ using a cost function $f$ defined by Equation~\ref{e.opf_f}, as follows:

\begin{eqnarray}
\label{e.opf_f}
f_{\max}(\langle s \rangle) & = & \left\{ \begin{array}{ll}
  0 & \mbox{if $s\in S$,} \\
+\infty & \mbox{otherwise}
\end{array}\right. \nonumber \\
f_{\max}(\pi_s \cdot \langle s,t \rangle) & = & \max\{f_{\max}(\pi_s),d(s,t)\}.
\end{eqnarray}
where $f_{\max}(\pi_s \cdot \langle s,t \rangle)$ is the maximum distance between adjacent samples along the path $\pi_s \cdot \langle s,t \rangle$. Thus, its training algorithm minimizes $f_{\max}$ for every sample $t \in \mathcal{Z}_1$, assigning an optimum-path $P(t)$ with a minimum cost defined by Equation~\ref{e.opf_cost}, as follows:

\begin{eqnarray}
\label{e.opf_cost}
C(t) = \min_{\forall \pi_t \in ({\mathcal Z}_1, {\mathcal A})}\{f_{\max}(\pi_t)\}.
\end{eqnarray}

Finally, Algorithm~\ref{a.opf_train} describes the $f_{\max}$ minimization procedure.

\begin{algorithm}[!h]
\KwIn{Training set $\mathcal{Z}_1$, 	prototypes $\mathcal{S}$, priority queue $Q$, current cost $c$.}
\KwOut{Optimum-path forest $P$ and cost map $C$.}
\For{$s \in \mathcal{Z}_1$}{
	$C(s) \leftarrow +\infty$\;
}
\For{$s \in \mathcal{S}$}{
	$C(s) \leftarrow 0$; $P(s) \leftarrow -1$; Insert $s$ in $Q$\;
}
\While{$Q$ is not empty}{
	Remove $s$ from $Q$ such that $C(s)$ is minimum\;
	\For{$t \in \mathcal{Z}_1$, such that $t \neq s$ and $C(t) > C(s)$}{
		$c \leftarrow \max\{C(s), d(s,t)\}$\;
		\If{$c < C(t)$}{
			\If{$C(t) \neq +\infty$}{
				Remove $t$ from $Q$\;
			}
			$C(t) \leftarrow c$; $P(t) \leftarrow s$; Insert $t$ in $Q$\;
		}
	}
}
\caption{OPF training algorithm.}
\label{a.opf_train} 
\end{algorithm}

\subsubsection{Testing Step}
\label{sss.opf_test}

During the testing phase, each sample $t$ will be connected to a sample $s \in \mathcal{V}_1$, becoming part of the original graph. The algorithm's goal is to find an optimum-path $P(t)$ that connects a prototype to node $t$, which is achieved by evaluation the path through an optimum-cost function denoted by Equation~\ref{e.opf_test}, as follows:

\begin{equation}
\label{e.opf_test}
C(t) = \min_{\forall s \in {\mathcal Z}_1}\{{\max \{C(s),d(s,t)\}}\}.\end{equation}

\subsection{Distance Measures}
\label{ss.dist}

Distance is a mathematical formulation used to describe closeness or farness between entities, i.e., provides a value that defines whether a pair of entities is similar or not. It is usually calculated between two vectors $x$ and $y$, where $d(x,y)$ is a function that defines the distance between both vectors as a non-negative real number. Nonetheless, a set of properties must be satisfied to consider $d(\cdot)$ a metric, as follows:

\begin{itemize}
	\item Identity: $d(x,y) = 0$ if $x = y$;
	\item Symmetry: $d(x,y) = d(y,x)$;
	\item Triangle Inequality: $d(x,y) \leq d(x,z) + d(z,y)$;
	\item Non-negativity: $d(x, y) \geq 0$.
\end{itemize}

This work follows the taxonomy proposed by Alfeilat et al.~\cite{Abu:19} and divides the distance functions into eight categories. Table~\ref{t.distance} describes the employed distance metrics, as well as their taxonomy and formula.

\renewcommand{\arraystretch}{1.275}
\begin{center}
	\footnotesize
	\begin{tabularx}{1.7\linewidth}
	 {>{\hsize=0.275\hsize}l
	  >{\hsize=0.19\hsize}l
	  >{\hsize=0.3\hsize}c} 
	  	\caption{Distance measures and their meta-information.}
        \\ \toprule \endfirsthead
		\toprule \endhead
		\midrule\multicolumn{3}{r}{\itshape continues on next page}\\\midrule\endfoot
	    \bottomrule\endlastfoot
		\textbf{Taxonomy} & \textbf{Distance} & \textbf{Formula}
		\\ \midrule
		\multirow{6}{*}{$L_p$} & Chebyshev ($D_1$) & $\max\limits_{i} |x_i-y_i|$
		\\
		& Chi-Squared ($D_2$) & $\sqrt{\sum_{i=1}^n \frac{(x_i-y_i)^2}{|x_i+y_i|}}$
		\\
		& Euclidean ($D_3$) & $\sqrt{\sum_{i=1}^n |x_i-y_i|^2}$
		\\
		& Gaussian ($D_4$) & $e^{-\sqrt{\sum_{i=1}^n (x_i-y_i)^2}}$
		\\
		& Log-Euclidean ($D_5$) & $\log(\sqrt{\sum_{i=1}^n |x_i-y_i|^2})$
		\\
		& Manhattan ($D_6$) & $\sum_{i=1}^n |x_i - y_i|$
		\\ \midrule
		\multirow{7}{*}{$L_1$} & Bray-Curtis ($D_7$) & $\frac{\sum_{i=1}^n |x_i-y_i|}{\sum_{i=1}^n(x_i+y_1)}$
		\\
		& Canberra ($D_8$) & $\sum_{i=1}^n \frac{|x_i-y_i|}{|x_i| + |y_i|}$
		\\ 
		& Gower ($D_9$) & $\frac{\sum_{i=1}^n |x_i-y_i|}{n}$
		\\
		& Kulczynski ($D_{10}$) & $\frac{\sum_{i=1}^n|x_i-y_i|}{\sum_{i=1}^n \min(x_i,y_i)}$
		\\
		& Lorentzian ($D_{11}$) & $\sum_{i=1}^n	 e^{1+|x_i-y_i|}$
		\\
		& Non-Intersection ($D_{12}$) & $\frac{1}{2} \sum_{i=1}^n |x_i-y_i|$
		\\
		& Soergel ($D_{13}$) & $\frac{\sum_{i=1}^n |x_i-y_i|}{\sum_{i=1}^n \max(x_i,y_i)}$
		\\ \midrule
		\multirow{4}{*}{Inner Product} & Chord ($D_{14}$) & $\sqrt{2 - 2 \frac{\sum_{i=1}^n x_iy_i}{\sum_{i=1}^n x_i^2	\sum_{i=1}^n y_i^2}}$
		\\
		& Cosine ($D_{15}$) & $1 - \frac{\sum_{i=1}^n x_iy_i}{\sum_{i=1}^n x_i^2 \sum_{i=1}^n y_i^2}$
		\\
		& Dice ($D_{16}$) & $1 - \frac{\sum_{i=1}^n x_iy_i}{\sum_{i=1}^n x_i^2 + \sum_{i=1}^n y_i^2}$
		\\
		& Jaccard ($D_{17}$) & $\frac{\sum_{i=1}^n (x_i-y_i)^2}{\sum_{i=1}^n x_i^2 + \sum_{i=1}^n y_i^2 - \sum_{i=1}^n x_iy_i}$
		\\ \midrule
		\multirow{4}{*}{Squared Chord} & Bhattacharyya ($D_{18}$) & $-e^{\sum_{i=1}^n \sqrt{x_iy_i}}$
		\\
		& Hellinger ($D_{19}$) & $\sqrt{2 \sum_{i=1}^n (\sqrt{x_i}-\sqrt{y_i})^2}$
		\\
		& Matusita ($D_{20}$) & $\sqrt{\sum_{i=1}^n (\sqrt{x_i}-\sqrt{y_i})^2}$
		\\
		& Squared Chord ($D_{21}$) & $\sum_{i=1}^n (\sqrt{x_i} - \sqrt{y_i})^2$
		\\ \midrule
		\multirow{11}{*}{Squared $L_2$} & Additive Symmetric $\mathcal{X}^2$ ($D_{22}$) & $2 \sum_{i=1}^n \frac{(x_i-y_i)^2(x_i+y_i)}{x_iy_i}$
		\\
		& Average Euclidean ($D_{23}$) & $\sqrt{\frac{1}{n} \sum_{i=1}^n (x_i-y_i)^2}$
		\\
		& Clark ($D_{24}$) & $\sqrt{\sum_{i=1}^n (\frac{x_i-y_i}{|x_i|+|y_i|})^2}$
		\\
		& Divergence ($D_{25}$) & $2 \sum_{i=1}^n \frac{(x_i-y_i)^2}{(x_i+y_i)^2}$
		\\
		& Log-Squared Euclidean ($D_{26}$) & $\log(\sum_{i=1}^n (x_i-y_i)^2)$
		\\
		& Mean Censored Euclidean ($D_{27}$) & $\frac{\sum_{i=1}^n (x_i-y_i)^2}{\sum_{i=1}^n 1_{x_i^2+y_i^2 \neq 0}}$
		\\
		& Neyman $\mathcal{X}^2$ ($D_{28}$) & $\sum_{i=1}^n \frac{(x_i-y_i)^2}{x_i}$
		\\
		& Pearson $\mathcal{X}^2$ ($D_{29}$) & $\sum_{i=1}^n \frac{(x_i-y_i)^2}{y_i}$
		\\
		& Sangvi $\mathcal{X}^2$ ($D_{30}$) & $2 \sum_{i=1}^n \frac{(x_i-y_i)^2}{x_i+y_i}$
		\\
		& Squared $\mathcal{X}^2$ ($D_{31}$) & $\sum_{i=1}^n \frac{(x_i-y_i)^2}{x_i+y_i}$
		\\
		& Squared Euclidean ($D_{32}$) & $\sum_{i=1}^n (x_i-y_i)^2$
		\\ \midrule
		\multirow{6}{*}{Shannon Entropy} & Jeffreys ($D_{33}$) & $\sum_{i=1}^n (x_i-y_i) e^{\frac{x_i}{y_i}}$
		\\
		& Jensen ($D_{34}$) & $\frac{1}{2}[\sum_{i=1}^n \frac{x_i e^{x_i} + y_i e^{y_i}}{2} - (\frac{x_i+y_i}{2})e^{(\frac{x_i+y_i}{2})}]$
		\\
		& Jensen-Shannon ($D_{35}$) & $\frac{1}{2}[\sum_{i=1}^n x_i e^{(\frac{2x_i}{x_i+y_i})} + \sum_{i=1}^n y_i e^{(\frac{2y_i}{x_i+y_i})}]$
		\\
		& K-Divergence ($D_{36}$) & $\sum_{i=1}^n x_i e^{\frac{2x_i}{x_i+y_i}}$
		\\
		& Kullback-Leibler ($D_{37}$) & $\sum_{i=1}^n x_i e^{\frac{x_i}{y_i}}$
		\\
		& Topsoe ($D_{38}$) & $\sum_{i=1}^n x_i e^{(\frac{2x_i}{x_i+y_i})} + \sum_{i=1}^n y_i e^{(\frac{2y_i}{x_i+y_i})}$
		\\ \midrule
		\multirow{6}{*}{Vicissitude} & Max Symmetric $\mathcal{X}^2$ ($D_{39}$) & $\max(\sum_{i=1}^n \frac{(x_i-y_i)^2}{x_i}, \sum_{i=1}^n \frac{(x_i-y_i)^2}{y_i})$
		\\
		& Min Symmetric $\mathcal{X}^2$ ($D_{40}$) & $\min(\sum_{i=1}^n \frac{(x_i-y_i)^2}{x_i}, \sum_{i=1}^n \frac{(x_i-y_i)^2}{y_i})$
		\\
		& Vicis Symmetric 1 ($D_{41}$) & $\sum_{i=1}^n \frac{(x_i-y_i)^2}{\min(x_i,y_i)^2}$
		\\
		& Vicis Symmetric 2 ($D_{42}$) & $\sum_{i=1}^n \frac{(x_i-y_i)^2}{\min(x_i,y_i)}$
		\\
		& Vicis Symmetric 3 ($D_{43}$) & $\sum_{i=1}^n \frac{(x_i-y_i)^2}{\max(x_i,y_i)}$
		\\
		& Vicis-Wave Hedges ($D_{44}$) & $\sum_{i=1}^n \frac{|x_i-y_i}{\min(x_i,y_i)}$
		\\ \midrule
		\multirow{4}{*}{Other} & Hamming ($D_{45}$) & $\sum_{i=1}^n 1_{x_i \neq y_i}$
		\\
		& \multirow{2}{*}{Hassanat ($D_{46}$)} & $\sum_{i=1}^n 1 - \frac{1+\min(x_i,y_i)}{1+\max(x_i,y_i)}$ if $\min(x_i,y_i) \geq 0$
		\\
		& & $\sum_{i=1}^n 1 - \frac{1+\min(x_i,y_i)+|\min(x_i,y_i)|}{1+\max(x_i,y_i)+|\min(x_i,y_i)|}$ if $\min(x_i,y_i) < 0$
		\\
		& $\mathcal{X}^2$ Statistic ($D_{47}$) & $\sum_{i=1}^n \frac{x_i-m_i}{m_i}$, where $m_i=\frac{x_i+y_i}{2}$
		\label{t.distance}
	\end{tabularx}
\end{center}
\section{Methodology}
\label{s.methodology}

This section presents a brief discussion regarding the proposed approach, the employed datasets and a more in-depth explanation of the experimental setup.

\subsection{Proposed Approach}
\label{ss.proposed}

Optimum-Path Forests are graph-based classifiers and depend on a distance measure to calculate the cost of arcs. Essentially, every pair of nodes has their distance measured and compared to the rest of the graph, where the best paths (minimum costs) are defined as the prototypes. Moreover, new nodes inserted in the graph also have their distance calculated against the pre-defined prototypes and are conquered by the prototype that offers the best path.

It is important to remark that a proper choice of distance measures might affect a specific problem's outcome. Hence, the proposed approach verifies the suitability of employing different distance measures regarding Optimum-Path Forest classification applied over various domains.

Let $\mathcal{D}$ be the set of available distances, $\mathcal{Z}_1$ and $\mathcal{Z}_2$ be the training and testing sets of a particular dataset, respectively, and $O$ be the OPF classifier. After training and evaluating the classifiers using distinct distances measures, the evaluation metrics will be assessed and compared amongst themselves, determining whether they are feasible. Algorithm~\ref{a.proposed} describes the pseudo-code of the proposed approach.

\begin{algorithm}[!h]
\KwIn{Training set $\mathcal{Z}_1$, 	testing set $\mathcal{Z}_2$, set of distances $\mathcal{D}$, OPF classifier $O$ and array of metrics $\mathbf{m}$.}
\KwOut{Matrix of metrics $M$.}
\For{$d \in \mathcal{D}$}{
	$O \leftarrow$ train over $\mathcal{Z}_1$ with distance $d$\;
	$\mathbf{m} \leftarrow$ evaluate $O$ over $\mathcal{Z}_2$\;
	$M \leftarrow [M, m]$\;
}
\caption{Proposed approach pseudo-code.}
\label{a.proposed} 
\end{algorithm}

\subsection{Datasets}
\label{ss.dataset}

The proposed approach aims at evaluating the Optimum-Path Forest in supervised classification tasks, currently implemented in the OPFython~\cite{Rosa:20} library. Table~\ref{t.datasets} describes an overview concerning the $22$ employed datasets, their task type, and their number of samples and features. Note that such a large amount of datasets attempt to overcome the diversity of problems, i.e., low/high number of features, small/large amounts of samples, and distinct domains.

\renewcommand{\arraystretch}{1.7}
\begin{center}
	\begin{tabularx}{1.7\linewidth}
	 {>{\hsize=0.275\hsize}l
	  >{\hsize=0.19\hsize}l
	  >{\hsize=0.3\hsize}c
	  >{\hsize=0.3\hsize}c} 
	  	\caption{Employed datasets used in the computations.}
        \\ \toprule \endfirsthead
		\toprule \endhead
		\midrule\multicolumn{3}{r}{\itshape continues on next page}\\\midrule\endfoot
	    \bottomrule\endlastfoot
		\textbf{Dataset} & \textbf{Task} & \textbf{Samples} & \textbf{Features}
		\\ \midrule
		Arcene & Mass Spectrometry & $200$ & $10,000$
		\\ \midrule
		BASEHOCK & Text & $1,993$ & $4,862$
		\\ \midrule
		Caltech101 & Image Silhouettes & $8,671$ & $784$
		\\ \midrule
		COIL20 & Face Image & $1,540$ & $1,024$
		\\ \midrule                 
		Isolet & Spoken Letter Recognition & $1,560$ & $617$
		\\ \midrule
		Lung & Biological & $203$ & $3,312$
		\\ \midrule
		Madelon & Artificial & $2,600$ & $500$
		\\ \midrule
		MPEG7 & Image Silhouettes & $1,400$ & $1,024$
		\\ \midrule
		MPEG7-BAS & Image Descriptor & $1,400$ & $180$
		\\ \midrule
		MPEG7-Fourier & Image Descriptor & $1,400$ & $126$
		\\ \midrule
		Mushrooms & Biological & $8,124$ & $112$
		\\ \midrule
		NTL-Commercial & Energy Theft & $4,952$ & $8$
		\\ \midrule
		NTL-Industrial & Energy Theft & $3,182$ & $8$
		\\ \midrule
		ORL & Face Image & $400$ & $1,024$
		\\ \midrule
		PCMAC & Text & $1,943$ & $3,289$
		\\ \midrule
		Phishing & Network Security & $11,055$ & $68$
		\\ \midrule
		Segment & Image Segmentation & $2,310$ & $19$
		\\ \midrule
		Semeion & Handwritten Digits & $1,593$ & $256$
		\\ \midrule
		Sonar & Signal & $208$ & $60$
		\\ \midrule
		Spambase & Network Security & $4,601$ & $48$
		\\ \midrule
		Vehicle & Image Silhouettes & $846$ & $18$
		\\ \midrule
		Wine & Chemical & $178$ & $13$
		\label{t.datasets}
	\end{tabularx}
\end{center}

\subsection{Experimental Setup}
\label{ss.setup}

The experiments\footnote{The source code is available at: \url{https://github.com/gugarosa/opf_distance}.} are conducted using a $2$-fold cross-validation procedure with $25$ runnings, i.e., each dataset is split into training and testing sets using a unique seed through $25$ times. After the split, distance-based OPF classifiers and Scikit-Learn~\cite{Pedregosa:12} classifiers (Decision Tree, Logistic Regression and Support Vector Machine)\footnote{These classifiers use the standard hyperparameters proposed by Scikit-Learn.} are trained using the training set and evaluated over the testing set, providing insightful metrics for further analysis, such as classification accuracy. Thus, the experimental tables are composed of the accuracies' mean and standard deviation values\footnote{Time is not assessed between OPFs as there are no significant differences between calculating distinct distances.}.

Furthermore, to provide more robust analysis, we conduct two statistical tests, as follows:

\begin{itemize}
	\item Wilcoxon Signed-Rank: The accuracy distributions ($25$ values) are compared in pairs for each dataset using a significance of $0.05$, hence, determining whether distinct distance-based OPFs are statistically similar or not for a particular dataset;
	\item Friedman with Nemenyi post hoc: All dataset's accuracy distributions ($22 \times 25$ values) are merged and evaluated according to the Friedman ($0.05$ significance) and Nemenyi post hoc tests. Such a procedure provides a rank-based analysis that illustrates whether specific distance-based OPFs are better or worse than others.
\end{itemize}
\section{Experimental Results}
\label{s.experiments}

Tables~\ref{t.experiment_a},~\ref{t.experiment_b},~\ref{t.experiment_c},~\ref{t.experiment_d} and~\ref{t.experiment_e} portray the mean accuracy and standard deviation values over the testing sets evaluated by the proposed classifiers. According to the Wilcoxon signed-rank test, the bolded cells are statistically equivalent, while the underlined ones depict the highest mean accuracy. Initially, when comparing OPF-based classifiers' performance against state-of-the-art classifiers, it is possible to observe that OPF could not achieve the best performance according to Wilcoxon signed-rank test in $7$ out of $22$ datasets: Caltech101, Isolet, PCMAC, Phishing, Semeion, Spambase, and Wine. On the other hand, OPF-based classifiers could achieve comparable performance in $8$ datasets: Arcene, BASEHOCK, COIL20, Lung, Madelon, MPEG7, ORL, and Vehicle while surpassed the compared classifiers in the remaining $7$ datasets: MPEG7-BAS, MPEG7-Fourier, Mushrooms, NTL-Commercial, NTL-Industrial, Segment, and Sonar.

\begin{sidewaystable}
    \centering
	\caption{Mean OPF accuracy and standard deviation values over testing sets evaluated by $D_1-D_{10}$ classifiers.}
	\vspace*{0.3cm}
    \label{t.experiment_a}
    \scalebox{0.69}{
	\begin{tabular}{lcccccccccc}
		\toprule
		& $\mathbf{D_1}$ & $\mathbf{D_2}$ & $\mathbf{D_3}$ & $\mathbf{D_4}$ & $\mathbf{D_5}$ & $\mathbf{D_6}$ & $\mathbf{D_7}$ & $\mathbf{D_8}$ & $\mathbf{D_9}$ & $\mathbf{D_{10}}$
		\\ \midrule
		Arcene & $0.6168 \pm 0.0525$ & $0.7688 \pm 0.0339$ & $0.4792 \pm 0.0635$ & $0.7691 \pm 0.0385$ & $0.7333 \pm 0.0357$ & $0.5693 \pm 0.0417$ & $\underline{\mathbf{0.7797 \pm 0.0299}}$ & $\mathbf{0.7763 \pm 0.0342}$ & $0.7309 \pm 0.0332$ & $\mathbf{0.7763 \pm 0.0342}$
		\\
		BASEHOCK & $0.5707 \pm 0.0455$ & $0.7602 \pm 0.0233$ & $0.7457 \pm 0.1419$ & $\mathbf{0.9146 \pm 0.0077}$ & $0.4999 \pm 0.0052$ & $0.7053 \pm 0.0211$ & $0.7179 \pm 0.0342$ & $0.9107 \pm 0.0086$ & $0.5080 \pm 0.0258$ & $0.9107 \pm 0.0086$
		\\
		Caltech101 & $0.3705 \pm 0.0217$ & $0.5438 \pm 0.0057$ & $0.0279 \pm 0.0342$ & $0.5478 \pm 0.0052$ & $0.4786 \pm 0.0102$ & $0.0448 \pm 0.0201$ & $0.5435 \pm 0.0058$ & $0.5483 \pm 0.0053$ & $0.5364 \pm 0.0075$ & $0.5483 \pm 0.0053$
		\\
		COIL20 & $0.7180 \pm 0.0361$ & $0.9401 \pm 0.0094$ & $0.0479 \pm 0.0035$ & $0.9473 \pm 0.0084$ & $0.8769 \pm 0.0154$ & $0.7523 \pm 0.0191$ & $0.9419 \pm 0.0087$ & $0.9353 \pm 0.0106$ & $0.8788 \pm 0.0116$ & $0.9353 \pm 0.0106$
		\\
		Isolet & $0.3005 \pm 0.0105$ & $0.7733 \pm 0.0126$ & $0.0377 \pm 0.0028$ & $0.7221 \pm 0.0113$ & $0.6860 \pm 0.0143$ & $0.5525 \pm 0.0158$ & $0.7539 \pm 0.0126$ & $0.7780 \pm 0.0128$ & $0.6814 \pm 0.0138$ & $0.7780 \pm 0.0128$
		\\
		Lung & $0.7482 \pm 0.2268$ & $\mathbf{0.9182 \pm 0.0199}$ & $0.3022 \pm 0.2903$ & $\mathbf{0.9163 \pm 0.0208}$ & $0.8920 \pm 0.0381$ & $0.8186 \pm 0.0533$ & $\mathbf{0.9135 \pm 0.0217}$ & $\mathbf{0.9163 \pm 0.0223}$ & $0.8635 \pm 0.0553$ & $\mathbf{0.9163 \pm 0.0223}$
		\\
		Madelon & $0.6334 \pm 0.0121$ & $\underline{\mathbf{0.6364 \pm 0.0118}}$ & $0.5004 \pm 0.0046$ & $0.6291 \pm 0.0102$ & $0.6311 \pm 0.0105$ & $0.5749 \pm 0.0093$ & $0.6329 \pm 0.0117$ & $\mathbf{0.6363 \pm 0.0114}$ & $\mathbf{0.6357 \pm 0.0117}$ & $\mathbf{0.6363 \pm 0.0114}$
		\\
		MPEG7 & $0.4403 \pm 0.0195$ & $\mathbf{0.6993 \pm 0.0168}$ & $0.0133 \pm 0.0024$ & $\underline{\mathbf{0.7028 \pm 0.0159}}$ & $0.5082 \pm 0.0340$ & $0.0195 \pm 0.0043$ & $\mathbf{0.6992 \pm 0.0167}$ & $0.6963 \pm 0.0161$ & $0.5996 \pm 0.0163$ & $0.6963 \pm 0.0161$
		\\
		MPEG7-BAS & $0.6732 \pm 0.0170$ & $0.6748 \pm 0.0169$ & $0.0138 \pm 0.0016$ & $\underline{\mathbf{0.6893 \pm 0.0186}}$ & $\mathbf{0.6882 \pm 0.0184}$ & $0.5983 \pm 0.0173$ & $0.6739 \pm 0.0173$ & $0.6722 \pm 0.0172$ & $0.6709 \pm 0.0176$ & $0.6724 \pm 0.0173$
		\\
		MPEG7-Fourier & $0.1825 \pm 0.0087$ & $\mathbf{0.3549 \pm 0.0151}$ & $0.0139 \pm 0.0015$ & $0.3086 \pm 0.0156$ & $0.1126 \pm 0.0075$ & $0.3445 \pm 0.0147$ & $0.2439 \pm 0.0113$ & $0.3157 \pm 0.0145$ & $0.1074 \pm 0.0075$ & $0.3157 \pm 0.0145$
		\\
		Mushrooms & $0.9448 \pm 0.0951$ & $0.9726 \pm 0.0472$ & $0.5049 \pm 0.0170$ & $0.9657 \pm 0.0916$ & $0.8240 \pm 0.1342$ & $0.5073 \pm 0.0161$ & $0.9619 \pm 0.0694$ & $0.9613 \pm 0.0801$ & $0.8408 \pm 0.1295$ & $0.9339 \pm 0.1097$
		\\
		NTL-Commercial & $0.9342 \pm 0.0039$ & $0.9101 \pm 0.0059$ & $0.9250 \pm 0.0049$ & $0.9139 \pm 0.0050$ & $\underline{\mathbf{0.9748 \pm 0.0039}}$ & $0.8600 \pm 0.0396$ & $0.9336 \pm 0.0036$ & $0.9430 \pm 0.0046$ & $0.9675 \pm 0.0043$ & $0.9430 \pm 0.0046$
		\\
		NTL-Industrial & $0.9345 \pm 0.0074$ & $0.9144 \pm 0.0061$ & $0.9254 \pm 0.0251$ & $0.9233 \pm 0.0072$ & $\underline{\mathbf{0.9733 \pm 0.0041}}$ & $0.8891 \pm 0.0439$ & $0.9348 \pm 0.0070$ & $0.9325 \pm 0.0052$ & $0.9652 \pm 0.0042$ & $0.9325 \pm 0.0052$
		\\
		ORL & $0.6104 \pm 0.0304$ & $0.6476 \pm 0.0352$ & $0.0237 \pm 0.0041$ & $\mathbf{0.6551 \pm 0.0362}$ & $\mathbf{0.6499 \pm 0.0366}$ & $0.2561 \pm 0.0257$ & $0.6387 \pm 0.0358$ & $0.6041 \pm 0.0326$ & $0.6192 \pm 0.0315$ & $0.6041 \pm 0.0326$
		\\
		PCMAC & $0.5450 \pm 0.0461$ & $0.6813 \pm 0.0274$ & $0.6535 \pm 0.0951$ & $0.8005 \pm 0.0118$ & $0.4989 \pm 0.0078$ & $0.6229 \pm 0.0169$ & $0.6588 \pm 0.0281$ & $0.7936 \pm 0.0151$ & $0.4982 \pm 0.0077$ & $0.7936 \pm 0.0151$
		\\
		Phishing & $0.8222 \pm 0.0659$ & $0.8835 \pm 0.0205$ & $0.5156 \pm 0.0550$ & $0.8898 \pm 0.0271$ & $0.9083 \pm 0.0169$ & $0.6364 \pm 0.0460$ & $0.8813 \pm 0.0298$ & $0.8805 \pm 0.0242$ & $0.9245 \pm 0.0080$ & $0.8712 \pm 0.0281$
		\\
		Segment & $0.9107 \pm 0.0105$ & $0.9337 \pm 0.0071$ & $0.1429 \pm 0.0039$ & $\mathbf{0.9429 \pm 0.0070}$ & $0.9159 \pm 0.0112$ & $0.9090 \pm 0.0079$ & $0.9272 \pm 0.0079$ & $0.9328 \pm 0.0074$ & $0.8869 \pm 0.0121$ & $0.9335 \pm 0.0075$
		\\
		Semeion & $0.5033 \pm 0.0597$ & $0.8356 \pm 0.0139$ & $0.0985 \pm 0.0040$ & $0.8461 \pm 0.0112$ & $0.7909 \pm 0.0234$ & $0.0997 \pm 0.0044$ & $0.8369 \pm 0.0152$ & $0.8416 \pm 0.0106$ & $0.8219 \pm 0.0122$ & $0.8416 \pm 0.0106$
		\\
		Sonar & $0.6651 \pm 0.0363$ & $0.7400 \pm 0.0431$ & $0.4862 \pm 0.0359$ & $0.7438 \pm 0.0470$ & $0.7369 \pm 0.0406$ & $0.7031 \pm 0.0480$ & $\mathbf{0.7518 \pm 0.0463}$ & $0.7269 \pm 0.0476$ & $0.7208 \pm 0.0426$ & $0.7269 \pm 0.0476$
		\\
		Spambase & $0.7567 \pm 0.0514$ & $0.7763 \pm 0.0555$ & $0.4810 \pm 0.0866$ & $0.8355 \pm 0.0436$ & $0.8636 \pm 0.0210$ & $0.5823 \pm 0.0990$ & $0.7899 \pm 0.0569$ & $0.8437 \pm 0.0180$ & $0.8651 \pm 0.0260$ & $0.8432 \pm 0.0459$
		\\
		Vehicle & $0.6068 \pm 0.0142$ & $0.6327 \pm 0.0235$ & $0.2489 \pm 0.0114$ & $0.6527 \pm 0.0201$ & $\mathbf{0.6642 \pm 0.0199}$ & $0.5912 \pm 0.0200$ & $0.6333 \pm 0.0190$ & $0.6372 \pm 0.0222$ & $0.6313 \pm 0.0157$ & $0.6372 \pm 0.0222$
		\\
		Wine & $0.9364 \pm 0.0203$ & $0.9269 \pm 0.0224$ & $0.3224 \pm 0.0676$ & $0.9125 \pm 0.0272$ & $0.9299 \pm 0.0228$ & $0.9012 \pm 0.0229$ & $0.9415 \pm 0.0206$ & $0.9072 \pm 0.0246$ & $0.9457 \pm 0.0243$ & $0.9072 \pm 0.0246$
		\\ \bottomrule
	\end{tabular}}
\end{sidewaystable}

\begin{sidewaystable}
    \centering
	\caption{Mean OPF accuracy and standard deviation values over testing sets evaluated by $D_{11}-D_{20}$ classifiers.}
	\vspace*{0.3cm}
    \label{t.experiment_b}
    \scalebox{0.69}{
	\begin{tabular}{lcccccccccc}
		\toprule
		& $\mathbf{D_{11}}$ & $\mathbf{D_{12}}$ & $\mathbf{D_{13}}$ & $\mathbf{D_{14}}$ & $\mathbf{D_{15}}$ & $\mathbf{D_{16}}$ & $\mathbf{D_{17}}$ & $\mathbf{D_{18}}$ & $\mathbf{D_{19}}$ & $\mathbf{D_{20}}$
		\\ \midrule
		Arcene & $\mathbf{0.7720 \pm 0.0327}$ & $0.7272 \pm 0.0390$ & $0.7688 \pm 0.0339$ & $0.4440 \pm 0.0763$ & $0.7667 \pm 0.0343$ & $0.6696 \pm 0.0280$ & $\mathbf{0.7715 \pm 0.0321}$ & $\mathbf{0.7792 \pm 0.0326}$ & $\mathbf{0.7720 \pm 0.0327}$ & $0.7603 \pm 0.0268$
		\\
		BASEHOCK & $0.8971 \pm 0.0105$ & $0.5136 \pm 0.0319$ & $0.7645 \pm 0.0229$ & $0.5009 \pm 0.0049$ & $0.6870 \pm 0.0348$ & $0.8258 \pm 0.0163$ & $0.6798 \pm 0.0335$ & $0.7297 \pm 0.0316$ & $0.8971 \pm 0.0105$ & $0.7320 \pm 0.0304$
		\\
		Caltech101 & $0.5478 \pm 0.0052$ & $0.5438 \pm 0.0070$ & $0.5439 \pm 0.0059$ & $0.0014 \pm 0.0009$ & $0.5437 \pm 0.0057$ & $0.2418 \pm 0.0131$ & $0.5442 \pm 0.0056$ & $0.5444 \pm 0.0060$ & $0.5478 \pm 0.0051$ & $0.5440 \pm 0.0059$
		\\
		COIL20 & $0.9381 \pm 0.0080$ & $0.8829 \pm 0.0103$ & $0.9401 \pm 0.0094$ & $0.0159 \pm 0.0056$ & $0.9522 \pm 0.0081$ & $0.6464 \pm 0.0169$ & $0.9537 \pm 0.0094$ & $0.9382 \pm 0.0093$ & $0.9381 \pm 0.0080$ & $0.9150 \pm 0.0110$
		\\
		Isolet & $0.7735 \pm 0.0114$ & $0.6826 \pm 0.0146$ & $0.7733 \pm 0.0126$ & $0.0025 \pm 0.0019$ & $0.7207 \pm 0.0123$ & $0.4382 \pm 0.0139$ & $0.7091 \pm 0.0128$ & $0.7473 \pm 0.0131$ & $0.7735 \pm 0.0114$ & $0.6072 \pm 0.0136$
		\\
		Lung & $\mathbf{0.9171 \pm 0.0225}$ & $0.8627 \pm 0.0542$ & $\mathbf{0.9182 \pm 0.0199}$ & $0.0065 \pm 0.0175$ & $0.9184 \pm 0.0165$ & $0.4716 \pm 0.0579$ & $0.9153 \pm 0.0173$ & $0.9085 \pm 0.0249$ & $\mathbf{0.9171 \pm 0.0225}$ & $0.8886 \pm 0.0296$
		\\
		Madelon & $\mathbf{0.6360 \pm 0.0112}$ & $\mathbf{0.6357 \pm 0.0117}$ & $\underline{\mathbf{0.6364 \pm 0.0118}}$ & $0.5065 \pm 0.0483$ & $0.6309 \pm 0.0110$ & $0.5003 \pm 0.0091$ & $0.6290 \pm 0.0109$ & $0.6331 \pm 0.0120$ & $\mathbf{0.6360 \pm 0.0112}$ & $0.6329 \pm 0.0120$
		\\
		MPEG7 & $\mathbf{0.7027 \pm 0.0159}$ & $0.6010 \pm 0.0214$ & $\mathbf{0.6993 \pm 0.0169}$ & $0.0041 \pm 0.0045$ & $\mathbf{0.6992 \pm 0.0171}$ & $0.1952 \pm 0.0206$ & $\mathbf{0.6994 \pm 0.0171}$ & $\mathbf{0.6990 \pm 0.0171}$ & $\mathbf{0.7027 \pm 0.0160}$ & $0.6977 \pm 0.0160$
		\\
		MPEG7-BAS & $0.6769 \pm 0.0174$ & $0.6709 \pm 0.0176$ & $0.6748 \pm 0.0169$ & $0.0095 \pm 0.0044$ & $0.6882 \pm 0.0184$ & $0.4789 \pm 0.0170$ & $\mathbf{0.6884 \pm 0.0182}$ & $0.6736 \pm 0.0171$ & $0.6769 \pm 0.0174$ & $0.6736 \pm 0.0171$
		\\
		MPEG7-Fourier & $\underline{\mathbf{0.3562 \pm 0.0144}}$ & $0.1074 \pm 0.0075$ & $\mathbf{0.3549 \pm 0.0151}$ & $0.0130 \pm 0.0025$ & $0.3032 \pm 0.0151$ & $0.0483 \pm 0.0055$ & $0.2866 \pm 0.0144$ & $0.2332 \pm 0.0114$ & $\underline{\mathbf{0.3562 \pm 0.0144}}$ & $0.2270 \pm 0.0112$
		\\
		Mushrooms & $0.9874 \pm 0.0184$ & $0.9096 \pm 0.0678$ & $0.9761 \pm 0.0396$ & $0.2073 \pm 0.0172$ & $0.9916 \pm 0.0220$ & $0.9862 \pm 0.0304$ & $0.9796 \pm 0.0369$ & $0.9994 \pm 0.0013$ & $0.9629 \pm 0.0622$ & $0.9957 \pm 0.0196$
		\\
		NTL-Commercial & $0.9100 \pm 0.0052$ & $0.9674 \pm 0.0043$ & $0.9101 \pm 0.0059$ & $0.9455 \pm 0.0017$ & $0.9142 \pm 0.0050$ & $0.8165 \pm 0.0592$ & $0.9145 \pm 0.0050$ & $0.9339 \pm 0.0035$ & $0.9100 \pm 0.0052$ & $0.9338 \pm 0.0036$
		\\
		NTL-Industrial & $0.9145 \pm 0.0062$ & $0.9652 \pm 0.0043$ & $0.9144 \pm 0.0061$ & $0.9370 \pm 0.0025$ & $0.9239 \pm 0.0075$ & $0.7853 \pm 0.0774$ & $0.9242 \pm 0.0076$ & $0.9349 \pm 0.0071$ & $0.9145 \pm 0.0062$ & $0.9345 \pm 0.0070$
		\\
		ORL & $0.6424 \pm 0.0333$ & $0.6192 \pm 0.0315$ & $0.6476 \pm 0.0352$ & $0.0011 \pm 0.0016$ & $\mathbf{0.6584 \pm 0.0337}$ & $0.5453 \pm 0.0325$ & $\mathbf{0.6588 \pm 0.0346}$ & $0.6352 \pm 0.0359$ & $0.6424 \pm 0.0333$ & $0.6317 \pm 0.0356$
		\\
		PCMAC & $0.7890 \pm 0.0111$ & $0.5009 \pm 0.0077$ & $0.6849 \pm 0.0278$ & $0.4993 \pm 0.0105$ & $0.6372 \pm 0.0286$ & $0.6992 \pm 0.0140$ & $0.6339 \pm 0.0325$ & $0.6690 \pm 0.0317$ & $0.7890 \pm 0.0111$ & $0.6736 \pm 0.0310$
		\\
		Phishing & $0.8851 \pm 0.0348$ & $0.9258 \pm 0.0210$ & $0.8819 \pm 0.0301$ & $0.2874 \pm 0.0221$ & $0.9207 \pm 0.0335$ & $0.8896 \pm 0.0132$ & $0.8922 \pm 0.0397$ & $0.9266 \pm 0.0040$ & $0.8885 \pm 0.0163$ & $0.9176 \pm 0.0166$
		\\
		Segment & $0.9339 \pm 0.0073$ & $0.8866 \pm 0.0110$ & $0.9337 \pm 0.0071$ & $0.0000 \pm 0.0000$ & $\underline{\mathbf{0.9435 \pm 0.0063}}$ & $0.4694 \pm 0.0127$ & $0.9403 \pm 0.0074$ & $0.9239 \pm 0.0088$ & $0.9339 \pm 0.0073$ & $0.9214 \pm 0.0089$
		\\
		Semeion & $0.8461 \pm 0.0112$ & $0.8321 \pm 0.0108$ & $0.8365 \pm 0.0156$ & $0.0148 \pm 0.0046$ & $0.8371 \pm 0.0141$ & $0.6725 \pm 0.0385$ & $0.8377 \pm 0.0154$ & $0.8376 \pm 0.0132$ & $0.8462 \pm 0.0112$ & $0.8376 \pm 0.0131$
		\\
		Sonar & $0.7331 \pm 0.0459$ & $0.7208 \pm 0.0426$ & $0.7400 \pm 0.0431$ & $0.4541 \pm 0.0447$ & $\mathbf{0.7567 \pm 0.0449}$ & $0.5051 \pm 0.0330$ & $\underline{\mathbf{0.7610 \pm 0.0434}}$ & $\mathbf{0.7490 \pm 0.0468}$ & $0.7331 \pm 0.0459$ & $0.7300 \pm 0.0472$
		\\
		Spambase & $0.8225 \pm 0.0235$ & $0.8435 \pm 0.0652$ & $0.7720 \pm 0.0515$ & $0.4812 \pm 0.0473$ & $0.8405 \pm 0.0167$ & $0.8106 \pm 0.0355$ & $0.7234 \pm 0.0728$ & $0.8351 \pm 0.0122$ & $0.8293 \pm 0.0330$ & $0.8261 \pm 0.0207$
		\\
		Vehicle & $0.6379 \pm 0.0224$ & $0.6308 \pm 0.0163$ & $0.6327 \pm 0.0235$ & $0.2942 \pm 0.0395$ & $0.6496 \pm 0.0177$ & $0.5528 \pm 0.0156$ & $0.6526 \pm 0.0186$ & $0.6262 \pm 0.0196$ & $0.6379 \pm 0.0224$ & $0.6171 \pm 0.0155$
		\\
		Wine & $0.9140 \pm 0.0252$ & $0.9457 \pm 0.0243$ & $0.9269 \pm 0.0224$ & $0.0391 \pm 0.0444$ & $0.9331 \pm 0.0216$ & $0.4961 \pm 0.0320$ & $0.9379 \pm 0.0212$ & $0.9436 \pm 0.0192$ & $0.9140 \pm 0.0252$ & $0.9397 \pm 0.0181$
		\\ \bottomrule
	\end{tabular}}
\end{sidewaystable}

\begin{sidewaystable}
    \centering
	\caption{Mean OPF accuracy and standard deviation values over testing sets evaluated by $D_{21}-D_{30}$ classifiers.}
	\vspace*{0.3cm}
    \label{t.experiment_c}
    \scalebox{0.69}{
	\begin{tabular}{lcccccccccc}
		\toprule
		& $\mathbf{D_{21}}$ & $\mathbf{D_{22}}$ & $\mathbf{D_{23}}$ & $\mathbf{D_{24}}$ & $\mathbf{D_{25}}$ & $\mathbf{D_{26}}$ & $\mathbf{D_{27}}$ & $\mathbf{D_{28}}$ & $\mathbf{D_{29}}$ & $\mathbf{D_{30}}$
		\\ \midrule
		Arcene & $\mathbf{0.7784 \pm 0.0325}$ & $\mathbf{0.7784 \pm 0.0325}$ & $0.5837 \pm 0.0477$ & $0.7691 \pm 0.0385$ & $0.6379 \pm 0.0637$ & $0.7688 \pm 0.0339$ & $0.7688 \pm 0.0339$ & $0.7677 \pm 0.0342$ & $0.7667 \pm 0.0343$ & $\mathbf{0.7792 \pm 0.0326}$
		\\
		BASEHOCK & $0.7250 \pm 0.0332$ & $0.7246 \pm 0.0330$ & $0.5527 \pm 0.0344$ & $\mathbf{0.9145 \pm 0.0078}$ & $0.5354 \pm 0.0344$ & $0.7605 \pm 0.0268$ & $0.7640 \pm 0.0214$ & $0.6784 \pm 0.0328$ & $0.6867 \pm 0.0339$ & $0.7299 \pm 0.0322$
		\\
		Caltech101 & $0.5441 \pm 0.0060$ & $0.5443 \pm 0.0061$ & $0.0515 \pm 0.0179$ & $0.5478 \pm 0.0051$ & $0.0496 \pm 0.0163$ & $0.5437 \pm 0.0058$ & $0.5438 \pm 0.0058$ & $0.5442 \pm 0.0058$ & $0.5439 \pm 0.0061$ & $0.5444 \pm 0.0060$
		\\
		COIL20 & $0.9410 \pm 0.0092$ & $0.9410 \pm 0.0092$ & $0.2025 \pm 0.0515$ & $0.9473 \pm 0.0084$ & $0.5406 \pm 0.0221$ & $0.9401 \pm 0.0094$ & $0.9401 \pm 0.0094$ & $\underline{\mathbf{0.9550 \pm 0.0087}}$ & $0.9522 \pm 0.0081$ & $0.9382 \pm 0.0093$
		\\
		Isolet & $0.7543 \pm 0.0125$ & $0.7543 \pm 0.0125$ & $0.1504 \pm 0.0324$ & $0.7221 \pm 0.0113$ & $0.2662 \pm 0.0301$ & $0.7733 \pm 0.0126$ & $0.7733 \pm 0.0126$ & $0.7067 \pm 0.0136$ & $0.7207 \pm 0.0123$ & $0.7473 \pm 0.0131$
		\\
		Lung & $0.9103 \pm 0.0238$ & $0.9103 \pm 0.0238$ & $0.6520 \pm 0.0948$ & $\mathbf{0.9163 \pm 0.0208}$ & $0.7265 \pm 0.1045$ & $\mathbf{0.9182 \pm 0.0199}$ & $\mathbf{0.9182 \pm 0.0199}$ & $\underline{\mathbf{0.9210 \pm 0.0162}}$ & $0.9184 \pm 0.0165$ & $0.9085 \pm 0.0249$
		\\
		Madelon & $0.6331 \pm 0.0118$ & $0.6331 \pm 0.0118$ & $0.5000 \pm 0.0116$ & $0.6291 \pm 0.0102$ & $0.5005 \pm 0.0132$ & $\underline{\mathbf{0.6364 \pm 0.0118}}$ & $\underline{\mathbf{0.6364 \pm 0.0118}}$ & $0.6283 \pm 0.0111$ & $0.6309 \pm 0.0110$ & $0.6331 \pm 0.0120$
		\\
		MPEG7 & $\mathbf{0.6989 \pm 0.0170}$ & $\mathbf{0.6989 \pm 0.0170}$ & $0.1685 \pm 0.0190$ & $\mathbf{0.7027 \pm 0.0159}$ & $0.1694 \pm 0.0173$ & $\mathbf{0.6992 \pm 0.0168}$ & $\mathbf{0.6993 \pm 0.0171}$ & $\mathbf{0.6991 \pm 0.0167}$ & $\mathbf{0.6992 \pm 0.0169}$ & $\mathbf{0.6990 \pm 0.0171}$
		\\
		MPEG7-BAS & $0.6738 \pm 0.0173$ & $0.6738 \pm 0.0173$ & $0.0306 \pm 0.0139$ & $\underline{\mathbf{0.6893 \pm 0.0186}}$ & $0.0459 \pm 0.0173$ & $0.6748 \pm 0.0169$ & $0.6748 \pm 0.0169$ & $\mathbf{0.6884 \pm 0.0184}$ & $0.6882 \pm 0.0184$ & $0.6736 \pm 0.0171$
		\\
		MPEG7-Fourier & $0.2376 \pm 0.0113$ & $0.2376 \pm 0.0113$ & $0.0216 \pm 0.0078$ & $0.3086 \pm 0.0156$ & $0.0229 \pm 0.0069$ & $\mathbf{0.3549 \pm 0.0151}$ & $\mathbf{0.3549 \pm 0.0151}$ & $0.3006 \pm 0.0148$ & $0.3032 \pm 0.0151$ & $0.2332 \pm 0.0114$
		\\
		Mushrooms & $0.9874 \pm 0.0614$ & $0.9931 \pm 0.0251$ & $0.9688 \pm 0.0876$ & $0.9629 \pm 0.0919$ & $0.9684 \pm 0.0758$ & $0.9685 \pm 0.0392$ & $0.9692 \pm 0.0594$ & $0.9836 \pm 0.0339$ & $0.9967 \pm 0.0143$ & $0.9994 \pm 0.0013$
		\\
		NTL-Commercial & $0.9338 \pm 0.0035$ & $0.9338 \pm 0.0035$ & $0.7996 \pm 0.1762$ & $0.9139 \pm 0.0050$ & $0.7908 \pm 0.1710$ & $0.9101 \pm 0.0059$ & $0.9101 \pm 0.0059$ & $0.9142 \pm 0.0049$ & $0.9142 \pm 0.0050$ & $0.9339 \pm 0.0035$
		\\
		NTL-Industrial & $0.9349 \pm 0.0073$ & $0.9349 \pm 0.0073$ & $0.7667 \pm 0.2302$ & $0.9233 \pm 0.0072$ & $0.7520 \pm 0.2358$ & $0.9144 \pm 0.0061$ & $0.9144 \pm 0.0061$ & $0.9239 \pm 0.0075$ & $0.9239 \pm 0.0075$ & $0.9349 \pm 0.0071$
		\\
		ORL & $0.6372 \pm 0.0358$ & $0.6372 \pm 0.0358$ & $0.0585 \pm 0.0204$ & $\mathbf{0.6551 \pm 0.0362}$ & $0.0667 \pm 0.0208$ & $0.6476 \pm 0.0352$ & $0.6476 \pm 0.0352$ & $\mathbf{0.6572 \pm 0.0337}$ & $\mathbf{0.6584 \pm 0.0337}$ & $0.6352 \pm 0.0359$
		\\
		PCMAC & $0.6661 \pm 0.0291$ & $0.6655 \pm 0.0291$ & $0.5518 \pm 0.0395$ & $0.8004 \pm 0.0121$ & $0.5409 \pm 0.0425$ & $0.6816 \pm 0.0276$ & $0.6815 \pm 0.0272$ & $0.6333 \pm 0.0308$ & $0.6382 \pm 0.0301$ & $0.6692 \pm 0.0299$
		\\
		Phishing & $0.9099 \pm 0.0292$ & $0.9210 \pm 0.0073$ & $0.9184 \pm 0.0088$ & $0.9282 \pm 0.0153$ & $0.9050 \pm 0.0221$ & $0.8851 \pm 0.0189$ & $0.8833 \pm 0.0227$ & $0.9004 \pm 0.0343$ & $0.9315 \pm 0.0151$ & $0.9266 \pm 0.0039$
		\\
		Segment & $0.9257 \pm 0.0086$ & $0.9257 \pm 0.0086$ & $0.2556 \pm 0.0539$ & $\mathbf{0.9429 \pm 0.0070}$ & $0.3137 \pm 0.0527$ & $0.9337 \pm 0.0071$ & $0.9337 \pm 0.0071$ & $\mathbf{0.9428 \pm 0.0056}$ & $\underline{\mathbf{0.9435 \pm 0.0063}}$ & $0.9239 \pm 0.0088$
		\\
		Semeion & $0.8373 \pm 0.0136$ & $0.8373 \pm 0.0135$ & $0.6075 \pm 0.0258$ & $0.8463 \pm 0.0111$ & $0.6185 \pm 0.0243$ & $0.8357 \pm 0.0138$ & $0.8359 \pm 0.0125$ & $0.8365 \pm 0.0131$ & $0.8381 \pm 0.0146$ & $0.8376 \pm 0.0134$
		\\
		Sonar & $\mathbf{0.7515 \pm 0.0453}$ & $\mathbf{0.7515 \pm 0.0453}$ & $0.5669 \pm 0.0843$ & $0.7438 \pm 0.0470$ & $0.5928 \pm 0.0739$ & $0.7400 \pm 0.0431$ & $0.7400 \pm 0.0431$ & $\mathbf{0.7587 \pm 0.0438}$ & $\mathbf{0.7567 \pm 0.0449}$ & $\mathbf{0.7490 \pm 0.0468}$
		\\
		Spambase & $0.8210 \pm 0.0263$ & $0.8264 \pm 0.0150$ & $0.5242 \pm 0.0869$ & $0.8450 \pm 0.0517$ & $0.5071 \pm 0.0774$ & $0.7585 \pm 0.0658$ & $0.7285 \pm 0.0860$ & $0.8167 \pm 0.0372$ & $0.8451 \pm 0.0440$ & $0.8396 \pm 0.0163$
		\\
		Vehicle & $0.6307 \pm 0.0194$ & $0.6307 \pm 0.0194$ & $0.2527 \pm 0.0557$ & $0.6527 \pm 0.0201$ & $0.2883 \pm 0.0538$ & $0.6327 \pm 0.0235$ & $0.6327 \pm 0.0235$ & $0.6514 \pm 0.0183$ & $0.6496 \pm 0.0177$ & $0.6262 \pm 0.0196$
		\\
		Wine & $0.9433 \pm 0.0193$ & $0.9433 \pm 0.0193$ & $0.5299 \pm 0.1004$ & $0.9125 \pm 0.0272$ & $0.6033 \pm 0.1295$ & $0.9269 \pm 0.0224$ & $0.9269 \pm 0.0224$ & $0.9364 \pm 0.0199$ & $0.9331 \pm 0.0216$ & $0.9436 \pm 0.0192$
		\\ \bottomrule
	\end{tabular}}
\end{sidewaystable}

\begin{sidewaystable}
    \centering
	\caption{Mean OPF accuracy and standard deviation values over testing sets evaluated by $D_{31}-D_{40}$ classifiers.}
	\vspace*{0.3cm}
    \label{t.experiment_d}
    \scalebox{0.69}{
	\begin{tabular}{lcccccccccc}
		\toprule
		& $\mathbf{D_{31}}$ & $\mathbf{D_{32}}$ & $\mathbf{D_{33}}$ & $\mathbf{D_{34}}$ & $\mathbf{D_{35}}$ & $\mathbf{D_{36}}$ & $\mathbf{D_{37}}$ & $\mathbf{D_{38}}$ & $\mathbf{D_{39}}$ & $\mathbf{D_{40}}$
		\\ \midrule
		Arcene & $0.6608 \pm 0.0308$ & $0.7688 \pm 0.0339$ & $0.6896 \pm 0.0388$ & $0.6771 \pm 0.0435$ & $0.7667 \pm 0.0343$ & $0.5896 \pm 0.0491$ & $\underline{\mathbf{0.7797 \pm 0.0299}}$ & $0.7691 \pm 0.0385$ & $\underline{\mathbf{0.7797 \pm 0.0299}}$ & $\mathbf{0.7792 \pm 0.0326}$
		\\
		BASEHOCK & $0.6724 \pm 0.0211$ & $0.7639 \pm 0.0225$ & $0.6365 \pm 0.0512$ & $0.8298 \pm 0.0238$ & $0.6873 \pm 0.0339$ & $0.5342 \pm 0.0353$ & $0.7182 \pm 0.0339$ & $\mathbf{0.9149 \pm 0.0076}$ & $0.7179 \pm 0.0327$ & $0.7298 \pm 0.0321$
		\\
		Caltech101 & $0.3836 \pm 0.0075$ & $0.5439 \pm 0.0060$ & $0.1714 \pm 0.0134$ & $0.2638 \pm 0.0125$ & $0.5438 \pm 0.0057$ & $0.1009 \pm 0.0164$ & $0.5439 \pm 0.0062$ & $0.5478 \pm 0.0052$ & $0.5441 \pm 0.0058$ & $0.5444 \pm 0.0060$
		\\
		COIL20 & $0.8608 \pm 0.0172$ & $0.9401 \pm 0.0094$ & $0.3123 \pm 0.0114$ & $0.1915 \pm 0.0203$ & $0.9522 \pm 0.0081$ & $0.5680 \pm 0.0254$ & $0.9419 \pm 0.0087$ & $0.9473 \pm 0.0084$ & $0.9419 \pm 0.0087$ & $0.9382 \pm 0.0093$
		\\
		Isolet & $0.2942 \pm 0.0086$ & $0.7733 \pm 0.0126$ & $0.4943 \pm 0.0140$ & $0.3872 \pm 0.0264$ & $0.7207 \pm 0.0123$ & $0.3516 \pm 0.0164$ & $0.7539 \pm 0.0126$ & $0.7221 \pm 0.0113$ & $0.7539 \pm 0.0126$ & $0.7473 \pm 0.0131$
		\\
		Lung & $0.8165 \pm 0.0500$ & $\mathbf{0.9182 \pm 0.0199}$ & $0.6424 \pm 0.2049$ & $0.6384 \pm 0.2026$ & $0.9184 \pm 0.0165$ & $0.7712 \pm 0.1080$ & $\mathbf{0.9135 \pm 0.0217}$ & $\mathbf{0.9163 \pm 0.0208}$ & $\mathbf{0.9135 \pm 0.0217}$ & $0.9085 \pm 0.0249$
		\\
		Madelon & $\mathbf{0.6352 \pm 0.0128}$ & $\underline{\mathbf{0.6364 \pm 0.0118}}$ & $0.6335 \pm 0.0121$ & $0.6331 \pm 0.0132$ & $0.6309 \pm 0.0110$ & $\mathbf{0.6355 \pm 0.0125}$ & $0.6329 \pm 0.0117$ & $0.6291 \pm 0.0102$ & $0.6329 \pm 0.0117$ & $0.6331 \pm 0.0120$
		\\
		MPEG7 & $0.4654 \pm 0.0175$ & $\mathbf{0.6992 \pm 0.0169}$ & $0.1947 \pm 0.0097$ & $0.2245 \pm 0.0212$ & $\mathbf{0.6992 \pm 0.0169}$ & $0.2107 \pm 0.0174$ & $0.6987 \pm 0.0169$ & $\underline{\mathbf{0.7028 \pm 0.0160}}$ & $0.6988 \pm 0.0171$ & $\mathbf{0.6989 \pm 0.0171}$
		\\
		MPEG7-BAS & $0.6739 \pm 0.0167$ & $0.6748 \pm 0.0169$ & $0.6725 \pm 0.0178$ & $0.6755 \pm 0.0167$ & $0.6882 \pm 0.0184$ & $0.6702 \pm 0.0180$ & $0.6739 \pm 0.0173$ & $\underline{\mathbf{0.6893 \pm 0.0186}}$ & $0.6739 \pm 0.0173$ & $0.6736 \pm 0.0171$
		\\
		MPEG7-Fourier & $0.1874 \pm 0.0064$ & $\mathbf{0.3549 \pm 0.0151}$ & $0.1562 \pm 0.0079$ & $0.1611 \pm 0.0096$ & $0.3032 \pm 0.0151$ & $0.1464 \pm 0.0094$ & $0.2439 \pm 0.0113$ & $0.3086 \pm 0.0156$ & $0.2439 \pm 0.0113$ & $0.2332 \pm 0.0114$
		\\
		Mushrooms & $0.9613 \pm 0.0852$ & $0.9429 \pm 0.0816$ & $0.9942 \pm 0.0185$ & $0.9660 \pm 0.0729$ & $0.9985 \pm 0.0058$ & $0.9636 \pm 0.0833$ & $0.9634 \pm 0.0786$ & $0.9992 \pm 0.0013$ & $0.9719 \pm 0.0569$ & $0.9994 \pm 0.0013$
		\\
		NTL-Commercial & $0.9353 \pm 0.0038$ & $0.9101 \pm 0.0059$ & $0.9321 \pm 0.0036$ & $0.8799 \pm 0.0823$ & $0.9142 \pm 0.0050$ & $0.9080 \pm 0.0905$ & $0.9336 \pm 0.0036$ & $0.9139 \pm 0.0050$ & $0.9336 \pm 0.0036$ & $0.9339 \pm 0.0035$
		\\
		NTL-Industrial & $0.9355 \pm 0.0068$ & $0.9144 \pm 0.0061$ & $0.9350 \pm 0.0061$ & $0.9175 \pm 0.0391$ & $0.9239 \pm 0.0075$ & $0.9042 \pm 0.0428$ & $0.9348 \pm 0.0070$ & $0.9233 \pm 0.0072$ & $0.9348 \pm 0.0070$ & $0.9349 \pm 0.0071$
		\\
		ORL & $0.5860 \pm 0.0316$ & $0.6476 \pm 0.0352$ & $0.6073 \pm 0.0350$ & $0.5536 \pm 0.0360$ & $\mathbf{0.6584 \pm 0.0337}$ & $0.6100 \pm 0.0345$ & $0.6387 \pm 0.0358$ & $\mathbf{0.6551 \pm 0.0362}$ & $0.6387 \pm 0.0358$ & $0.6352 \pm 0.0359$
		\\
		PCMAC & $0.6330 \pm 0.0224$ & $0.6838 \pm 0.0292$ & $0.6124 \pm 0.0393$ & $0.6848 \pm 0.0787$ & $0.6377 \pm 0.0313$ & $0.5310 \pm 0.0371$ & $0.6604 \pm 0.0275$ & $0.8005 \pm 0.0120$ & $0.6603 \pm 0.0272$ & $0.6694 \pm 0.0299$
		\\
		Phishing & $0.8778 \pm 0.0527$ & $0.8869 \pm 0.0243$ & $0.9136 \pm 0.0056$ & $0.9097 \pm 0.0104$ & $0.9300 \pm 0.0364$ & $0.8876 \pm 0.0477$ & $0.9107 \pm 0.0214$ & $0.9198 \pm 0.0468$ & $0.8947 \pm 0.0384$ & $0.9264 \pm 0.0038$
		\\
		Segment & $0.9110 \pm 0.0092$ & $0.9337 \pm 0.0071$ & $0.9322 \pm 0.0079$ & $0.9120 \pm 0.0113$ & $\underline{\mathbf{0.9435 \pm 0.0063}}$ & $0.9155 \pm 0.0130$ & $0.9272 \pm 0.0079$ & $\mathbf{0.9429 \pm 0.0070}$ & $0.9272 \pm 0.0079$ & $0.9239 \pm 0.0088$
		\\
		Semeion & $0.6461 \pm 0.0217$ & $0.8356 \pm 0.0155$ & $0.6443 \pm 0.0322$ & $0.6187 \pm 0.0252$ & $0.8383 \pm 0.0134$ & $0.6165 \pm 0.0302$ & $0.8374 \pm 0.0130$ & $0.8463 \pm 0.0111$ & $0.8380 \pm 0.0138$ & $0.8375 \pm 0.0134$
		\\
		Sonar & $0.6659 \pm 0.0357$ & $0.7400 \pm 0.0431$ & $0.6931 \pm 0.0450$ & $0.6533 \pm 0.0502$ & $\mathbf{0.7567 \pm 0.0449}$ & $0.6662 \pm 0.0389$ & $\mathbf{0.7518 \pm 0.0463}$ & $0.7438 \pm 0.0470$ & $\mathbf{0.7518 \pm 0.0463}$ & $\mathbf{0.7490 \pm 0.0468}$
		\\
		Spambase & $0.8013 \pm 0.0233$ & $0.7666 \pm 0.0570$ & $0.8150 \pm 0.0397$ & $0.8153 \pm 0.0250$ & $0.8585 \pm 0.0151$ & $0.7771 \pm 0.0444$ & $0.7883 \pm 0.0457$ & $0.8579 \pm 0.0425$ & $0.8126 \pm 0.0371$ & $0.8398 \pm 0.0168$
		\\
		Vehicle & $0.6013 \pm 0.0174$ & $0.6327 \pm 0.0235$ & $0.6440 \pm 0.0215$ & $0.6246 \pm 0.0201$ & $0.6496 \pm 0.0177$ & $0.6129 \pm 0.0177$ & $0.6333 \pm 0.0190$ & $0.6527 \pm 0.0201$ & $0.6333 \pm 0.0190$ & $0.6262 \pm 0.0196$
		\\
		Wine & $0.9394 \pm 0.0184$ & $0.9269 \pm 0.0224$ & $0.9293 \pm 0.0202$ & $0.8845 \pm 0.0331$ & $0.9331 \pm 0.0216$ & $0.9301 \pm 0.0482$ & $0.9415 \pm 0.0206$ & $0.9125 \pm 0.0272$ & $0.9415 \pm 0.0206$ & $0.9436 \pm 0.0192$
		\\ \bottomrule
	\end{tabular}}
\end{sidewaystable}

\begin{sidewaystable}
    \centering
	\caption{Mean OPF accuracy and standard deviation values over testing sets evaluated by $D_{41}-D_{47}$, DT, LR and SVM classifiers.}
	\vspace*{0.3cm}
    \label{t.experiment_e}
    \scalebox{0.685}{
	\begin{tabular}{lcccccccccc}
		\toprule
		& $\mathbf{D_{41}}$ & $\mathbf{D_{42}}$ & $\mathbf{D_{43}}$ & $\mathbf{D_{44}}$ & $\mathbf{D_{45}}$ & $\mathbf{D_{46}}$ & $\mathbf{D_{47}}$ & \textbf{DT} & \textbf{LR} & \textbf{SVM}
		\\ \midrule
		Arcene & $0.7688 \pm 0.0339$ & $0.4085 \pm 0.0656$ & $\mathbf{0.7784 \pm 0.0325}$ & $0.6619 \pm 0.0350$ & $0.7184 \pm 0.0394$ & $\mathbf{0.7784 \pm 0.0309}$ & $0.7320 \pm 0.0456$ & $0.6336 \pm 0.0479$ & $\mathbf{0.7755 \pm 0.0471}$ & $0.6323 \pm 0.0343$
		\\
		BASEHOCK & $0.7642 \pm 0.0229$ & $0.4960 \pm 0.0132$ & $0.7244 \pm 0.0328$ & $0.5602 \pm 0.0386$ & $0.6216 \pm 0.0359$ & $0.7102 \pm 0.0337$ & $0.5770 \pm 0.0409$ & $0.8926 \pm 0.0162$ & $0.7008 \pm 0.1225$ & $\underline{\mathbf{0.9217 \pm 0.0199}}$
		\\
		Caltech101 & $0.5441 \pm 0.0059$ & $0.0205 \pm 0.0112$ & $0.5443 \pm 0.0060$ & $0.1852 \pm 0.0194$ & $0.4034 \pm 0.0077$ & $0.5438 \pm 0.0059$ & $0.4039 \pm 0.0075$ & $0.3720 \pm 0.0076$ & $\underline{\mathbf{0.5570 \pm 0.0047}}$ & $0.5283 \pm 0.0080$
		\\
		COIL20 & $0.9401 \pm 0.0094$ & $0.0484 \pm 0.0069$ & $0.9410 \pm 0.0092$ & $0.8657 \pm 0.0148$ & $0.8602 \pm 0.0179$ & $0.9451 \pm 0.0088$ & $0.9034 \pm 0.0107$ & $0.8123 \pm 0.0242$ & $\mathbf{0.9545 \pm 0.0082}$ & $0.9143 \pm 0.0152$
		\\
		Isolet & $0.7733 \pm 0.0126$ & $0.0467 \pm 0.0092$ & $0.7543 \pm 0.0125$ & $0.2931 \pm 0.0109$ & $0.3001 \pm 0.0104$ & $0.7486 \pm 0.0117$ & $0.2959 \pm 0.0123$ & $0.6652 \pm 0.0261$ & $\underline{\mathbf{0.9148 \pm 0.0119}}$ & $0.8294 \pm 0.0230$
		\\
		Lung & $\mathbf{0.9182 \pm 0.0199}$ & $0.3911 \pm 0.1849$ & $0.9103 \pm 0.0238$ & $0.8324 \pm 0.0457$ & $0.8714 \pm 0.0352$ & $\mathbf{0.9174 \pm 0.0172}$ & $0.8586 \pm 0.0345$ & $0.7234 \pm 0.0558$ & $\mathbf{0.9145 \pm 0.0280}$ & $0.7859 \pm 0.0467$
		\\
		Madelon & $\underline{\mathbf{0.6364 \pm 0.0118}}$ & $0.4963 \pm 0.0128$ & $0.6331 \pm 0.0118$ & $\mathbf{0.6340 \pm 0.0112}$ & $\mathbf{0.6338 \pm 0.0118}$ & $\mathbf{0.6345 \pm 0.0113}$ & $\mathbf{0.6354 \pm 0.0093}$ & $\mathbf{0.6242 \pm 0.0356}$ & $0.5873 \pm 0.0110$ & $\mathbf{0.6310 \pm 0.0098}$
		\\
		MPEG7 & $\mathbf{0.6990 \pm 0.0168}$ & $0.0111 \pm 0.0049$ & $\mathbf{0.6990 \pm 0.0170}$ & $0.2995 \pm 0.0185$ & $0.5054 \pm 0.0174$ & $\mathbf{0.6989 \pm 0.0169}$ & $0.5047 \pm 0.0181$ & $0.4376 \pm 0.0228$ & $\mathbf{0.6975 \pm 0.0155}$ & $0.5306 \pm 0.0311$
		\\
		MPEG7-BAS & $0.6748 \pm 0.0169$ & $0.0254 \pm 0.0104$ & $0.6738 \pm 0.0173$ & $0.6682 \pm 0.0174$ & $0.6711 \pm 0.0169$ & $0.6752 \pm 0.0172$ & $0.6866 \pm 0.0182$ & $0.3559 \pm 0.0210$ & $0.2112 \pm 0.0281$ & $0.4530 \pm 0.0319$
		\\
		MPEG7-Fourier & $\mathbf{0.3549 \pm 0.0151}$ & $0.0331 \pm 0.0094$ & $0.2376 \pm 0.0113$ & $0.0911 \pm 0.0058$ & $0.1720 \pm 0.0068$ & $0.2614 \pm 0.0119$ & $0.0993 \pm 0.0064$ & $0.1076 \pm 0.0129$ & $0.0240 \pm 0.0114$ & $0.0323 \pm 0.0094$
		\\
		Mushrooms & $0.9906 \pm 0.0336$ & $0.3690 \pm 0.1202$ & $0.9997 \pm 0.0006$ & $\underline{\mathbf{0.9999 \pm 0.0003}}$ & $\mathbf{0.9997 \pm 0.0012}$ & $0.9815 \pm 0.0431$ & $\mathbf{0.9997 \pm 0.0012}$ & $0.9993 \pm 0.0011$ & $0.9989 \pm 0.0006$ & $0.9993 \pm 0.0006$
		\\
		NTL-Commercial & $0.9101 \pm 0.0059$ & $0.6985 \pm 0.2875$ & $0.9338 \pm 0.0035$ & $0.9662 \pm 0.0042$ & $0.9355 \pm 0.0041$ & $0.9326 \pm 0.0035$ & $0.9717 \pm 0.0040$ & $0.9483 \pm 0.0070$ & $0.9458 \pm 0.0017$ & $0.9457 \pm 0.0017$
		\\
		NTL-Industrial & $0.9144 \pm 0.0061$ & $0.7525 \pm 0.2384$ & $0.9349 \pm 0.0073$ & $0.9606 \pm 0.0043$ & $0.9359 \pm 0.0074$ & $0.9339 \pm 0.0063$ & $0.9684 \pm 0.0049$ & $0.9502 \pm 0.0139$ & $0.9384 \pm 0.0021$ & $0.9405 \pm 0.0023$
		\\
		ORL & $0.6476 \pm 0.0352$ & $0.0508 \pm 0.0181$ & $0.6372 \pm 0.0358$ & $0.5039 \pm 0.0310$ & $0.5887 \pm 0.0278$ & $\mathbf{0.6507 \pm 0.0360}$ & $0.6220 \pm 0.0314$ & $0.2843 \pm 0.0344$ & $\underline{\mathbf{0.6599 \pm 0.0382}}$ & $0.3395 \pm 0.0373$
		\\
		PCMAC & $0.6848 \pm 0.0261$ & $0.4906 \pm 0.0246$ & $0.6661 \pm 0.0284$ & $0.5527 \pm 0.0388$ & $0.6015 \pm 0.0337$ & $0.6536 \pm 0.0264$ & $0.5733 \pm 0.0394$ & $0.8332 \pm 0.0221$ & $0.5761 \pm 0.1002$ & $\underline{\mathbf{0.8443 \pm 0.0181}}$
		\\
		Phishing & $0.9076 \pm 0.0175$ & $0.3312 \pm 0.0458$ & $0.9143 \pm 0.0216$ & $0.9370 \pm 0.0067$ & $0.9305 \pm 0.0067$ & $0.9099 \pm 0.0187$ & $0.9301 \pm 0.0062$ & $0.9347 \pm 0.0034$ & $0.9372 \pm 0.0025$ & $\underline{\mathbf{0.9488 \pm 0.0024}}$
		\\
		Segment & $0.9337 \pm 0.0071$ & $0.1788 \pm 0.0305$ & $0.9257 \pm 0.0086$ & $0.6377 \pm 0.0239$ & $0.9120 \pm 0.0092$ & $0.9310 \pm 0.0086$ & $0.8160 \pm 0.0237$ & $0.9344 \pm 0.0067$ & $0.8869 \pm 0.0075$ & $0.9176 \pm 0.0062$
		\\
		Semeion & $0.8357 \pm 0.0139$ & $0.0386 \pm 0.0222$ & $0.8372 \pm 0.0134$ & $0.6614 \pm 0.0191$ & $0.7143 \pm 0.0173$ & $0.8370 \pm 0.0130$ & $0.7145 \pm 0.0182$ & $0.6410 \pm 0.0249$ & $0.8885 \pm 0.0106$ & $\underline{\mathbf{0.9088 \pm 0.0105}}$
		\\
		Sonar & $0.7400 \pm 0.0431$ & $0.5656 \pm 0.0943$ & $\mathbf{0.7515 \pm 0.0453}$ & $0.6282 \pm 0.0328$ & $0.6664 \pm 0.0379$ & $\mathbf{0.7541 \pm 0.0414}$ & $0.6633 \pm 0.0312$ & $0.6841 \pm 0.0269$ & $0.7195 \pm 0.0397$ & $0.7203 \pm 0.0546$
		\\
		Spambase & $0.7777 \pm 0.0615$ & $0.5220 \pm 0.0707$ & $0.8353 \pm 0.0177$ & $0.7893 \pm 0.0276$ & $0.8069 \pm 0.0215$ & $0.8158 \pm 0.0481$ & $0.8037 \pm 0.0261$ & $0.8838 \pm 0.0072$ & $0.9172 \pm 0.0032$ & $\underline{\mathbf{0.9205 \pm 0.0050}}$
		\\
		Vehicle & $0.6327 \pm 0.0235$ & $0.2610 \pm 0.0595$ & $0.6307 \pm 0.0194$ & $0.5941 \pm 0.0215$ & $0.6062 \pm 0.0159$ & $0.6397 \pm 0.0221$ & $0.6174 \pm 0.0181$ & $0.6439 \pm 0.0239$ & $0.6318 \pm 0.0204$ & $\underline{\mathbf{0.6768 \pm 0.0200}}$
		\\
		Wine & $0.9269 \pm 0.0224$ & $0.3490 \pm 0.0988$ & $0.9433 \pm 0.0193$ & $0.9376 \pm 0.0174$ & $0.9367 \pm 0.0209$ & $0.9403 \pm 0.0199$ & $0.9310 \pm 0.0211$ & $0.8537 \pm 0.0395$ & $0.9570 \pm 0.0231$ & $\underline{\mathbf{0.9696 \pm 0.0139}}$
		\\ \bottomrule
	\end{tabular}}
\end{sidewaystable}

Regarding the worst OPF-based classifiers performance, it is interesting to remark that in the Caltech101 dataset, $D_8$ and $D_{10}$ ($54.83\%$) were able to outperform DT ($37.20\%$) and SVM ($52.83\%$) but could not compete with LR ($55.70\%$) accuracy rates. In the Isolet dataset, $D_8$ and $D_{10}$ ($77.80\%$) could only surpass DT ($66.52\%$) and achieved a significant inferior result than SVM ($82.94\%$) and LR ($91.48\%$). Considering the PCMAC dataset, $D_{38}$ was the best OPF-based classifier and surpassed LR with a $39\%$ performance boost, yet it roughly achieved $96\%$ and $94\%$ of LR's and SVM's performance, respectively. In the Phishing dataset, $D_{44}$ ($93.70\%$) outperformed DT ($93.43\%$) however was slightly inferior to LR ($93.72\%$) and SVM ($94.88\%$). On the Semeion dataset, $D_{24}$ and $D_{38}$ ($84.63\%$) performances were better than DT ($64.10\%$) and worse than LR ($88.85\%$) and SVM ($90.88\%$). In the Spambase dataset, $D_{9}$ achieved around $98\%$, $93\%$, and $94\%$ of DT's, LR's, and SVM's performance, respectively. Finally, regarding the Wine dataset, $D_{9}$ and $D_{12}$ ($94.57\%$) bettered DT ($85.37\%$) and was outdone by LR ($95.70\%$) and SVM ($96.96\%$).

Considering the comparable OPF-based classifiers' performance, one can perceive that OPF achieved the highest mean accuracy in $5$ out of $7$ datasets: Arcene, COIL20, Lung, Madelon, and MPEG7, yet they were statistically equivalent to the state-of-the-art classifiers. In such datasets, it is possible to observe that OPF-based classifiers achieved between $0.5\%$ and $5\%$ more accuracy than the compared classifiers. Another point that should be highlighted is that every standard deviation appears to have two- and three-decimal precision, showing the robustness of the proposed approaches when employed with dissimilar data splits.

Regarding the best OPF-based classifiers performance, one can recognize that $D_{4}$ and $D_{38}$ had an outstanding accuracy ($68.93\%$) over the MPEG7-BAS dataset, achieving more than a $50\%$ improvement when compared to the best non-OPF classifier (SVM - $45.30\%$), while $D_{11}$ and $D_{19}$ ($35.62\%$) had an improvement of more than $200\%$ over DT in the MPEG7-Fourier dataset, which was the best non-OPF classifier ($10.76\%$). On the other hand, concerning the Mushrooms dataset, OPF-based performance increase was not that significant, i.e., slightly better than non-OPF classifiers ($99.99\%$ against $99.93\%$). Furthermore, concerning NTL datasets (Commercial and Industrial), $D_{5}$ managed to improve the non-OPF based classifiers' performance by approximately $3\%$. Regarding the Segment dataset, $D_{15}$, $D_{29}$, and $D_{35}$ achieved an accuracy rate of $94.35\%$, which was somewhat better than DT ($93.44\%$), SVM ($91.76\%$) and LR ($88.69\%$). Finally, $D_{17}$ ($76.10\%$) could outperform the non-based OPF classifiers in the Sonar dataset, e.g., DT ($68.41\%$), LR ($71.95\%$) and SVM ($72.03\%$).

\begin{sidewaysfigure}
	\centering
	\includegraphics[scale=0.6]{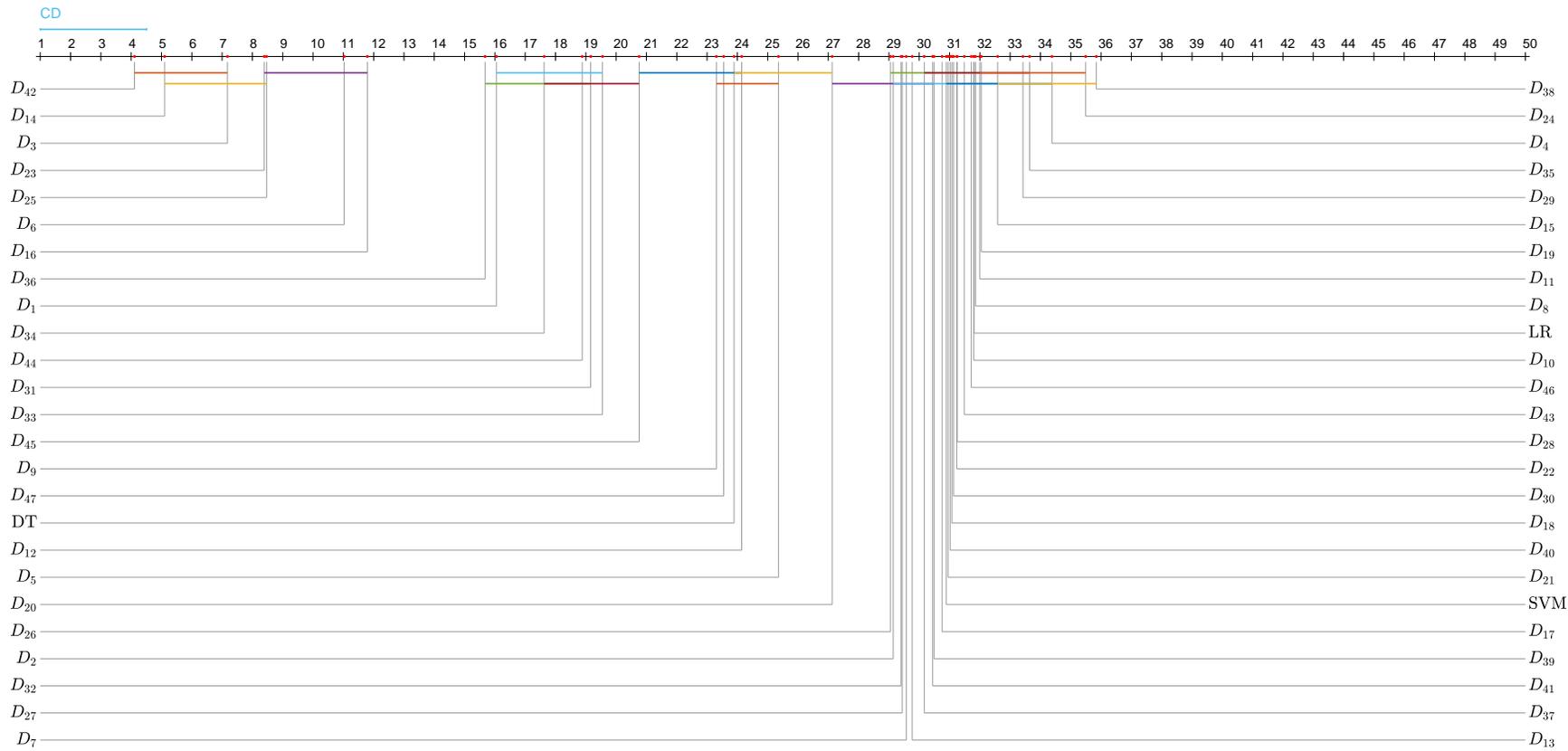}	
	\caption{Classifiers' ranking according to Friedman with Nemenyi post hoc statistical test.}
	\label{f.friedman_nemenyi}
\end{sidewaysfigure}

Apart from verifying whether OPF-based classifiers were better than non-OPF classifiers, we provided an additional statistical test capable of ranking individual classifiers' performance. Such a test stood for the Friedman statistical test with $5\%$ significance and followed by a Nemenyi post hoc. Figure~\ref{f.friedman_nemenyi} depicts individual classifiers ranking, and their critical difference calculated by Friedman's test, i.e., rank $1$ stands for the worst classifier while $50$ stands for the best one.

According to Figure~\ref{f.friedman_nemenyi}, it is possible to highlight that $D_{38}$ was the best classifier among all datasets and runnings, while $D_{42}$ was the worst one. Tables~\ref{t.experiment_d} and~\ref{t.experiment_e} confirm such a hypothesis, where $D_{38}$ achieved the best statistical equivalence in six datasets while having the highest mean accuracy in two out of those six datasets. On the other hand, $D_{42}$ had poor performance compared to any other classifier and did not achieve any best statistical equivalence nor the highest mean accuracy. Another remarkable point is that, despite their statistical equivalence, 9 OPF-based classifiers ($D_{38}, D_{24}, D_{4}, D_{35}, D_{29}, D_{15}, D_{19}, D_{11}, D_{8}$) were better ranked than the best non-OPF classifier (LR). Additionally, when ranking non-OPF based classifiers, LR could outperform both SVM and DT, where the latter achieved the worst rank.

Additionally, we analyzed the influence of distance measures according to the feature space size, where the datasets described in Table~\ref{t.datasets} were categorized by their feature size: low (up to $100$ features), medium (between $100$ and $1,000$ features), high ($1,000$ and more features). When working with low-dimensional spaces, it is possible to observe that $D_5$, $D_{15}$ and $D_{17}$ achieved the best results between distance measures in those datasets, yet according to Figure~\ref{f.friedman_nemenyi} only $D_{15}$ and $D_{17}$ had a reasonable performance across all datasets. Additional ``inner-product`` measures, such as $D_{14}$ and $D_{16}$, also had a better impact than $D_5$ considering all datasets, thus turning out in a good measure category for low-dimensional spaces.

Regarding medium-dimensional spaces, some $L_p$-based measures, such as $D_2$, $D_4$, $D_{11}$, and $D_{13}$, were able to achieve competitive results in those datasets, especially the $L_1$ ones, which attained the best performance when compared across the mentioned distances. Finally, considering the high-dimensional spaces, $D_7$, $D_8$, $D_{10}$, $D_{37}$, and $D_{39}$ achieved comparable results to the state-of-the-art classifier SVM. Apart from one ``Shannon-entropy`` and one ``vicissitude` distance, the remaining ones are $L_1$-based and had better performances according to Figure~\ref{f.friedman_nemenyi}. Such findings reinforce the capacity of $L_p$-based distances and strengthen one of the most used distances in OPF (log-euclidean). On the other hand, they provide insightful sources of whether distance measures directly impacts the performance of OPF classifiers or not.

\begin{figure}[!ht]
	\centering
	\includegraphics[scale=0.6]{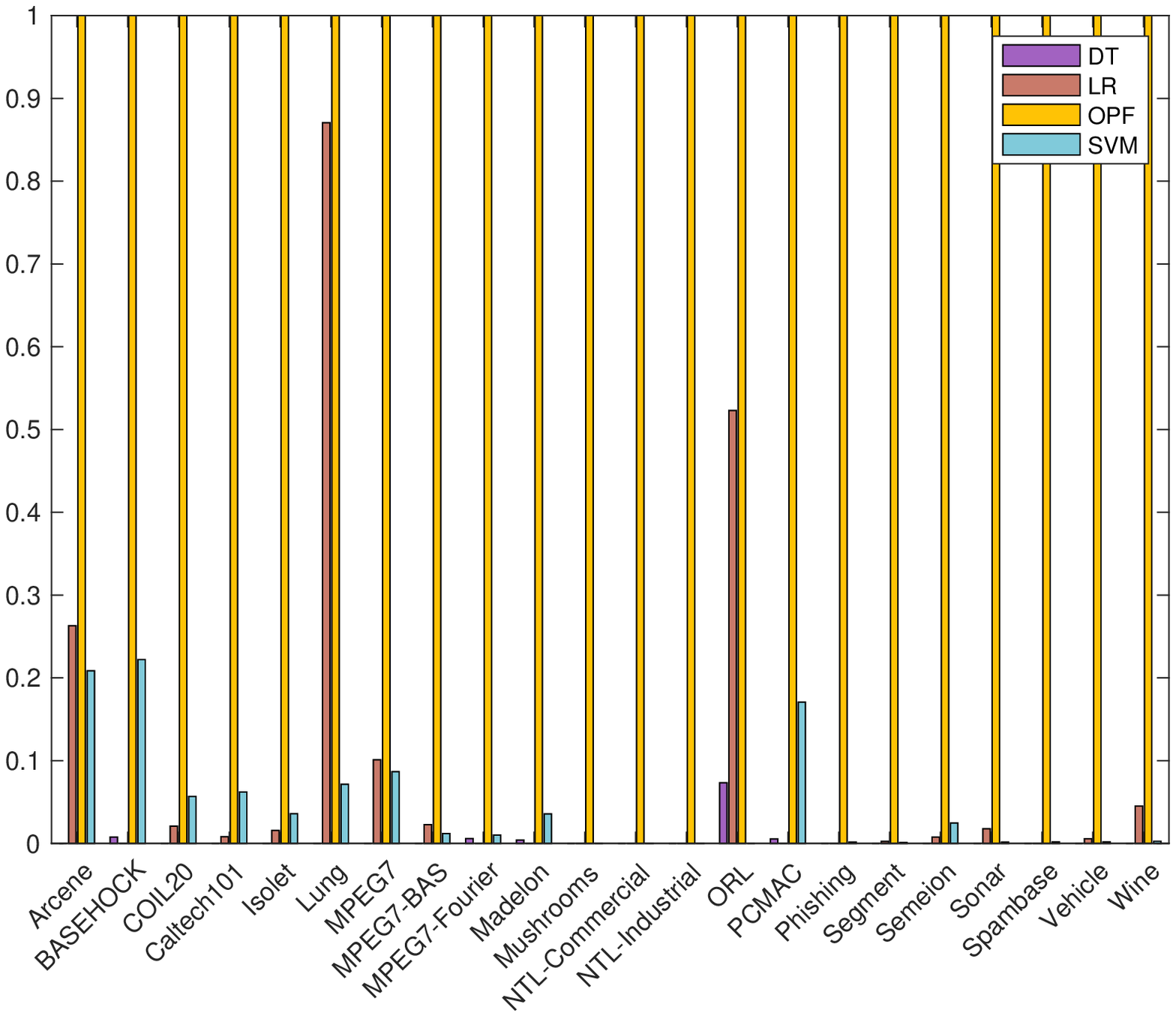}	
	\caption{Normalized execution time (training + testing) comparison between DT, LR, OPF and SVM classifiers.}
	\label{f.execution_time}
\end{figure}

Figure~\ref{f.execution_time} illustrates a vertical bar plot that compares the normalized execution times between DT, LR, OPF, and SVM. Note that we have just used a standard OPF classifier, without distance variation, as the time difference between distance metrics is unfeasible compared to OPF's running time. In such a figure, one can perceive that OPF had the longest-running time considering all datasets, followed by LR, SVM, and DT. Thus, one can conclude that even though OPF-based classifiers furnish performance improvements, they also have a higher computational burden, mainly because it has to calculate the distance between all nodes in the graph on both training and testing phases.

\section{Conclusion}
\label{s.conclusion}

This work performed a comparative study between $47$ distance measures applied to the OPF classifier considering supervised learning over $22$ datasets. Additionally, it compared OPF-based classifiers with three state-of-the-art classifiers, such as Decision Tree, Logistic Regression, and Support Vector Machine.

In most circumstances, OPF-based classifiers could outperform the baseline classifiers (DT, LR, and SVM), achieving statistically equivalent results according to the Wilcoxon signed-rank test with $5\%$ significance, and, in some cases, higher accuracy rates. Additionally, a statistical analysis through the Friedman-Nemenyi test was conducted to rank individual classifiers' performance over the experiments, showing that 9 OPF-based classifiers were better ranked than the best non-OPF classifier.

Even though OPF-based classifiers achieved outstanding results, they had a higher computational burden due to OPF's methodology, where it calculates distance measures between all nodes in the graph, elevating the algorithm's complexity. Nevertheless, it should be remarked that such a computational burden increase enabled OPF-based classifiers to outperform non-OPF classifiers, achieving over than $200\%$ boost in a couple of datasets.

Regarding future works, we aim at exploring alternatives to OPF's distance calculation, e.g., confidence and similarity measures. We believe that it is possible to pre-calculate the distance according to a more representative measure, such as a similarity score calculated by a Siamese Network, and further improve the generalization capacity of OPF-based classifiers.

\section*{Acknowledgments}
The authors are grateful to São Paulo Research Foundation (FAPESP) grants \#2013/07375-0, \#2014/12236-1, \#2019/07665-4, \#2019/02205-5, and \#2020/12101-0, and to the Brazilian National Council for Research and Development (CNPq) \#307066/2017-7 and \#427968/2018-6.

\bibliographystyle{unsrt}
\bibliography{paper}

\end{document}